\documentclass[runningheads]{llncs}

\usepackage[T1]{fontenc}
\usepackage{amsmath,amssymb,amsfonts}
\usepackage{mathtools}
\usepackage{booktabs}
\usepackage{multirow}
\usepackage{graphicx}
\usepackage{float}
\usepackage{xcolor}
\usepackage{url}
\usepackage{microtype}
\usepackage{enumitem}
\usepackage{algorithm}
\usepackage{algorithmic}
\usepackage{tikz}
\usepackage{pgfplots}
\pgfplotsset{compat=1.18}
\usetikzlibrary{positioning,arrows.meta,shapes.geometric,fit,backgrounds,calc}
\usepgfplotslibrary{groupplots}
\usepackage{colortbl}
\usepackage{pifont}

\definecolor{accentred}{RGB}{180,30,30}
\newcommand{\rev}[1]{#1}

\newcommand{\Letter}{\ding{41}} 
\newcommand{\rcog}{R_{\text{cog}}}
\newcommand{\reng}{R_{\text{eng}}}
\newcommand{\rdir}{R_{\text{dir}}}
\newcommand{\rtotal}{R_{\text{total}}}
\newcommand{\kquery}{\mathcal{K}_{\text{query}}}
\newcommand{\SE}{\textsc{SE}}
\newcommand{\CD}{\textsc{CD}}
\newcommand{\modelname}{\textsc{HeuristicEdu}}
\newcommand{\dataname}{\textsc{SocraticEdu}}
\newcommand{\sbertcite}{\cite{reimers2019sentencebert}}
\newcommand{\sbertcitefn}{\cite{reimers2019sentencebert}}

\begin{document}

\title{Beyond Direct Answering: Aligning Educational LLMs as
       Socratic Guides via Heuristic Reinforcement Learning}

\titlerunning{Beyond Direct Answering}

\author{Xiaokun Wang\inst{1,2} \and
        Siyu Song\inst{1} \and
        Wentao Liu\inst{1,2}\textsuperscript{(\Letter)} \and
        Xiaodong Zou\inst{3,4}\textsuperscript{(\Letter)}}
\authorrunning{X. Wang et al.}
\institute{East China Normal University, Shanghai, China \and
           Shanghai Chuangjie Situo Information Technology Co., Ltd.,
           Shanghai, China \and
           Shanghai Normal University, Shanghai, China \and
           China Association of STEM Education, China\\
           \textsuperscript{(\Letter)}\email{businesssituo@163.com; stem2603@126.com}}

\maketitle

\begin{abstract}
\rev{Large language models (LLMs) deployed in educational settings often
behave as \emph{direct answerers}: they disclose target concepts in the
opening turn instead of guiding students through progressive inquiry, as
Socratic pedagogy prescribes.}
\rev{We present {\modelname}, a two-phase pipeline that aligns
Qwen2.5-7B toward Socratic tutoring via supervised warm-up and Group
Relative Policy Optimization (GRPO).}
\rev{Training uses {\dataname}, 797 multi-turn Chinese children's science
dialogues reconstructed from a live platform, with a heuristic reward over
cognitive depth ($\rcog$), curiosity engagement ($\reng$), and directness
($\rdir$), together with a $\kquery$ correction for student-introduced terms.}
\rev{We introduce Scaffolding Effectiveness ({\SE}) and Conversation Depth
({\CD}) to evaluate outcomes beyond surface fluency.}
\rev{On 30 held-out questions, the best GRPO variant improves {\SE} from
30.0\% to 63.3\% and lowers keyword leakage from 30.0\% to 13.3\%.
Notably, this best variant omits the directness penalty during optimization,
suggesting that explicit anti-leakage terms can conflict with gradient-based
behavioral alignment.}
\rev{An unaligned Qwen-72B baseline reaches 0\% {\SE} and 96.7\% leakage,
showing that scale alone does not induce Socratic behavior.}
\end{abstract}

\section{Introduction}
\label{sec:intro}

Educational psychology has long distinguished between \emph{telling}
and \emph{teaching}: a student who is handed the answer to ``Why is
the sky blue?'' acquires a fact, whereas one led to discover Rayleigh
scattering through progressive questioning develops durable conceptual
understanding~\cite{vygotsky1978,bloom1956}.
Despite this, current LLMs deployed in K-12 platforms default to
comprehensive, immediate answers.
An analysis of 10,512 student queries from a live science platform
shows that every tested model---from a 7B instruction-tuned baseline
to Qwen-72B---leaks core scientific terminology in the opening
response at rates above 96\%, irrespective of model size.
This \textbf{Direct Answerer} failure mode is therefore behavioral
rather than capability-driven, and it cannot be resolved by scaling
alone.

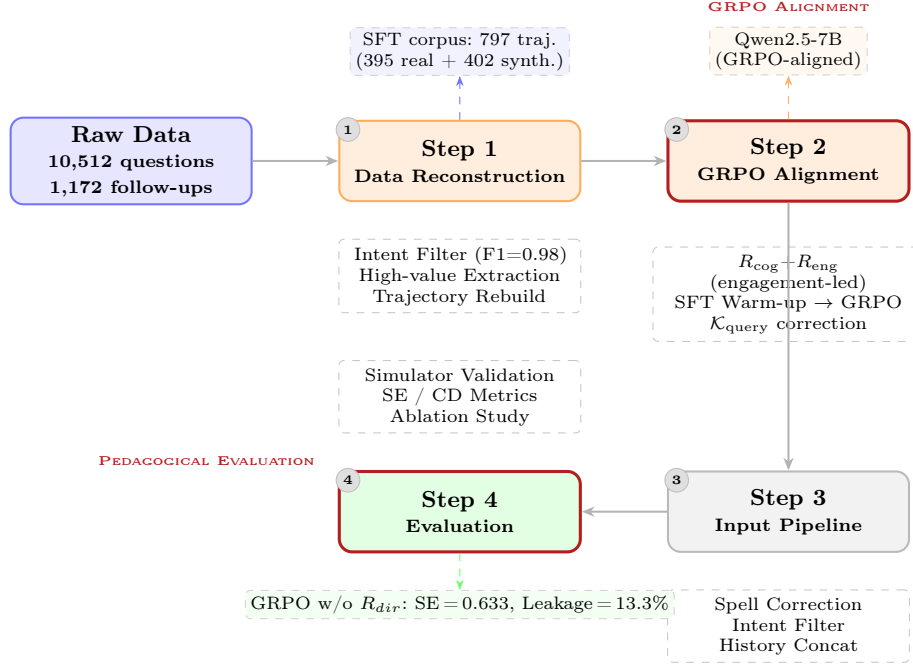
\begin{figure}[t]
\centering
\resizebox{\textwidth}{!}{%
\begin{tikzpicture}[
  box/.style    = {rectangle, rounded corners=5pt, draw, thick,
                   minimum width=33mm, minimum height=11mm,
                   align=center, font=\small\bfseries},
  databox/.style= {box, fill=blue!10, draw=blue!55},
  stepbox/.style= {box, fill=orange!13, draw=orange!65},
  pipeBox/.style= {box, fill=gray!10,  draw=gray!55},
  evalbox/.style= {box, fill=green!11, draw=green!55},
  corebox/.style= {very thick, draw=accentred},
  sub/.style    = {rectangle, rounded corners=2pt, draw=gray!45, dashed,
                   font=\scriptsize, fill=white!95, align=center,
                   inner sep=3pt, text width=31mm},
  ann/.style    = {rectangle, rounded corners=2pt, draw=gray!35, dashed,
                   font=\scriptsize, fill=white, align=center, inner sep=2pt},
  arr/.style    = {-{Stealth[length=5pt,width=4pt]}, thick, gray!60},
  darr/.style   = {-{Stealth[length=4pt,width=3pt]}, dashed, thin},
]

\node[databox] (raw) at (0,0)
  {\textbf{Raw Data}\\[-1pt]
   \scriptsize 10,512 questions\\[-1pt]\scriptsize 1,172 follow-ups};
\node[stepbox] (s1) at (4.5,0)
  {\textbf{Step 1}\\[-1pt]\scriptsize Data Reconstruction};
\node[stepbox, corebox] (s2) at (9,0)
  {\textbf{Step 2}\\[-1pt]\scriptsize GRPO Alignment};
\node[pipeBox] (s3) at (9,-4.8)
  {\textbf{Step 3}\\[-1pt]\scriptsize Input Pipeline};
\node[evalbox, corebox] (s4) at (4.5,-4.8)
  {\textbf{Step 4}\\[-1pt]\scriptsize Evaluation};

\node[sub, below=5mm of s1]
  {Intent Filter (F1=0.98)\\High-value Extraction\\Trajectory Rebuild};
\node[sub, below=6mm of s2, text width=35mm]
  {$\rcog$+$\reng$ (engagement-led)\\SFT Warm-up $\to$ GRPO\\$\kquery$ correction};
\node[sub, below=5mm of s3]
  {Spell Correction\\Intent Filter\\History Concat};
\node[sub, above=5mm of s4]
  {Simulator Validation\\SE / CD Metrics\\Ablation Study};

\draw[arr] (raw) -- (s1);
\draw[arr] (s1)  -- (s2);
\draw[arr] (s2.south) -- (s3.north);
\draw[arr] (s3)  -- (s4);

\node[ann, above=6mm of s1, fill=blue!5]
  (corpus) {\scriptsize SFT corpus: 797 traj.\\(395 real + 402 synth.)};
\draw[darr, blue!55] (s1.north) -- (corpus.south);

\node[ann, above=6mm of s2, fill=orange!5]
  (model) {\scriptsize Qwen2.5-7B\\(GRPO-aligned)};
\draw[darr, orange!60] (s2.north) -- (model.south);

\node[ann, below=5mm of s4, fill=green!5]
  (result) {\scriptsize GRPO w/o $R_{dir}$: SE\,=\,0.633, Leakage\,=\,13.3\%};
\draw[darr, green!60] (s4.south) -- (result.north);

\node[font=\tiny\bfseries\color{accentred}, above=1mm of model]
  {\textsc{GRPO Alignment}};
\node[font=\tiny\bfseries\color{accentred}, left=2mm of s4, yshift=7mm]
  {\textsc{Pedagogical Evaluation}};

\foreach \n/\x/\y in {1/4.5/0, 2/9/0, 3/9/-4.8, 4/4.5/-4.8}
  \node[circle, fill=gray!25, draw=gray!50, inner sep=1.5pt,
        font=\tiny\bfseries, anchor=north west]
    at (\x-1.65, \y+0.55) {\n};

\end{tikzpicture}
}
\caption{Overview of the {\modelname} pipeline: data reconstruction,
  GRPO alignment, inference-time preprocessing, and pedagogical evaluation
  with {\SE} and {\CD}.}
\label{fig:pipeline}
\end{figure}

\rev{To address this failure mode, we present {\modelname}, a pipeline for
aligning Qwen2.5-7B-Instruct toward Socratic tutoring (Figure~\ref{fig:pipeline}).
We reconstruct platform logs into {\dataname}, a 797-trajectory corpus spanning
seven elementary-science domains, then apply SFT followed by GRPO with a
heuristic reward over cognitive lift ($\rcog$), curiosity engagement ($\reng$),
and directness ($\rdir$), plus a $\kquery$ correction for student-introduced
terms.}
\rev{We evaluate pedagogical quality with outcome-oriented metrics in a four-turn
LLM-simulated tutoring loop, scoring whether the child articulates the target
concept without tutor-side keyword leakage.}

\rev{Our main contributions are summarized as follows:}
\begin{itemize}[leftmargin=1.5em, topsep=2pt, itemsep=4pt, parsep=0pt]
\item \rev{\textbf{Dataset.}
  We introduce {\dataname}, a 797-trajectory Socratic tutoring corpus
  reconstructed from live student--platform interactions and style-controlled
  synthesis across seven elementary-science domains.}

\item \rev{\textbf{Methodology.}
  We propose an SFT-then-GRPO alignment pipeline with heuristic rewards for
  cognitive depth, curiosity engagement, and controlled answer disclosure,
  including a $\kquery$ correction for student-introduced terms.}

\item \rev{\textbf{Evaluation and Findings.}
  We introduce {\SE} and {\CD} for closed-loop Socratic evaluation.
  On 30 held-out questions, PT-GRPO without $\rdir$ reaches {\SE}${}=0.633$ with
  13.3\% leakage, outperforming SFT-only, prompt-engineering, and unaligned
  Qwen-72B baselines; $\rdir$ helps as an inference-time constraint but hurts
  when used as a GRPO penalty.}
\end{itemize}

\section{Related Work}
\label{sec:related}

\paragraph{LLMs in Education.}
Intelligent tutoring systems built on dialogue have a long research
history~\cite{graesser2001autotutor}.
Recent LLM-based approaches demonstrate strong pedagogical
capabilities in mathematics and reading
comprehension~\cite{macina2023mathdial,tack2022ai,wei2022emergent,openai2023gpt4},
yet these systems predominantly operate in an answer-delivery mode
rather than adopting the guided-inquiry strategy advocated by
Socratic pedagogy.
The specific problem of aligning an LLM's \emph{response behavior}
toward Socratic scaffolding---rather than improving answer
correctness---has received little formal treatment.

\paragraph{Alignment and Pedagogical Reward Signals.}
RLHF~\cite{ouyang2022training} and process reward models~\cite{lightman2023lets}
align LLM behavior with preference or step-level feedback, but collecting
pedagogical preference pairs is costly.
We therefore adopt GRPO~\cite{shao2024deepseekmath}, which normalizes rewards
within sampled groups and avoids a learned critic.
For the cognitive component of the reward, Bloom's taxonomy
\cite{bloom1956,anderson2001taxonomy} provides a theory-grounded scale of
educational demand, which we operationalize with an LLM judge.

\paragraph{Beyond Surface-Level Evaluation.}
BLEU~\cite{papineni2002bleu} and ROUGE~\cite{lin2004rouge} correlate
poorly with pedagogical quality in open-domain tutoring, where
multiple valid responses exist for every student query.
Reference-free evaluation methods~\cite{mehri2020usr} are better
suited in principle, but they still mainly judge response quality
rather than whether a tutoring dialogue changes the learner's state.
Recent educational-dialogue work therefore argues for broader
benchmarks that assess adaptive guidance, cognitive growth, learner
autonomy, and conceptual progression~\cite{ilkou2025hybrid,liu2025guideeval}.
Outside education, progression-aware dialogue modeling similarly
evaluates whether a conversation moves toward a desired
outcome~\cite{sanders2022progression}.
Our \SE{} and \CD{} metrics build on this direction but instantiate it
for Socratic tutoring: they ask whether the student reaches the target
concept through self-expression and whether the dialogue continues to
advance semantically rather than repeat surface wording.

\section{The \dataname{} Dataset}
\label{sec:data}

\subsection{Data Source}
\dataname{} is constructed from interaction logs of \textbf{Dr.\
Curious}, a deployed educational Q\&A platform serving Chinese
primary-school students (ages 8--12) across 14 schools.
\rev{Students submit free-form science questions in conversational Chinese; the
platform returns an initial AI answer and supports follow-ups, yielding both
single-turn queries and multi-turn chains.
Each record contains the question text, timestamps, anonymized student/school
identifiers, view counts, and follow-up utterances.
No personally identifiable information is released with {\dataname}.}
Table~\ref{tab:data} summarizes the corpus statistics after each curation stage.

\begin{table}[H]
\caption{Corpus statistics for \dataname{}.}
\label{tab:data}
\centering\small
\begin{tabular}{lrr}
\toprule
\textbf{Split} & \textbf{Count} & \textbf{Notes} \\
\midrule
Raw questions          & 10,512 & All subjects \\
Real follow-up records &  1,172 & Multi-turn chains \\
\midrule
After intent filtering &  8,911 & Scientific only \\
High-value seeds       &    134 & Views $\geq 30$, inquiry patterns \\
Real SFT trajectories  &    395 & Reconstructed chains \\
Synthetic SFT traj.    &    402 & API-generated (3 styles) \\
\textbf{Total SFT corpus} & \textbf{797} & \\
\bottomrule
\end{tabular}
\end{table}

\subsection{Scientific Intent Filtering}
\rev{Because the platform accepts arbitrary free-form questions, we first train a
lightweight \emph{scientific-intent} filter.
Queries are represented with character-level TF--IDF features and classified by
logistic regression into \emph{scientific} vs.\ \emph{non-scientific} labels.
Five-fold cross-validation gives macro-F1 $=0.9805 \pm 0.0022$; applying the
classifier removes 1,601 off-topic items and retains 8,911 science-oriented
queries for clustering and trajectory reconstruction.}

\subsection{Clustering and Trajectory Construction}
\rev{We embed filtered questions with multilingual Sentence-BERT~\sbertcitefn{}
and apply K-Means clustering ($K{=}15$) to organize heterogeneous topics.
Clusters with above-median view count and inquiry-pattern density are treated as
\emph{high-engagement} themes, from which we select 134 seeds with at least 30
views and sustained follow-up behavior.}
\rev{For seeds with observed multi-turn logs, we reconstruct 395 \emph{real}
trajectories through dialogue replay, using logged student follow-ups when
available.
To cover seeds without long real chains, we further generate 402 \emph{synthetic}
trajectories under guiding, exploratory, and analogical tutoring styles.
The resulting 797-trajectory corpus spans seven semantic domains and provides
the supervised warm-up data before GRPO.}

\section{Heuristic Reward Function}
\label{sec:method}

We define the composite training reward as:
\begin{equation}
  \rtotal = \alpha\,\rcog + \beta\,\reng - \gamma\,\rdir
  \label{eq:rtotal}
\end{equation}
where $\alpha, \beta, \gamma \geq 0$ control the relative weight of
each component.  Section~\ref{sec:results} shows that $\gamma = 0$
is optimal in practice.

\paragraph{$\rcog$: Cognitive Depth.}
\rev{$\rcog$ encourages the tutor to raise the cognitive demand of the exchange
rather than merely confirm facts.
A Qwen-turbo judge independently assigns Bloom's Taxonomy levels
$b \in \{1,\ldots,6\}$ (remember $\to$ create) to the preceding student query
$q_{t-1}$ and the current model response $r_t$.}
\rev{The reward measures the normalized cognitive \emph{lift} between turns:}
\begin{equation}
  \rcog = \frac{b_{\text{response}} - b_{\text{query}}}{5}
  \in [-1,\,1]
\end{equation}
A positive value indicates that the response moves the student toward
higher-order thinking.

\paragraph{$\reng$: Curiosity Engagement.}
\rev{Socratic tutoring must sustain curiosity even when answers are withheld.
We therefore score each tutor turn $r_t$ with an Engagement Proxy Model (EPM)
that detects surface cues correlated with child follow-up in our logs, including
rhetorical questions, analogical framing, extreme-value language, and
open-ended hypothesis starters.}
\rev{Formally,}
\begin{equation}
  \reng = \textsc{EPM}(r_t) \in [0,\,1]
\end{equation}

\paragraph{$\rdir$: Directness Penalty with $\kquery$ Correction.}
\rev{The directness term discourages premature disclosure of the target concept.
For each seed question $i$, we curate a keyword set $\mathcal{K}_i$ containing
core scientific terms that a Socratic tutor should help the student discover
rather than state outright (e.g., \emph{photosynthesis} for a plant-growth
question).}
\rev{Let $\kquery$ denote terms in $\mathcal{K}_i$ already present in the
student's preceding utterance $q_{t-1}$.
We penalize only \emph{novel} leakage---terms in $r_t$ that were not introduced
by the student:}
\begin{equation}
  \rdir =
  \frac{|\{k \in \mathcal{K}_i \cap r_t\} \setminus \kquery|}
       {|\mathcal{K}_i|}
  \label{eq:rdir}
\end{equation}
Without this correction, a response that uses ``chlorophyll'' after a
student has asked ``Why does photosynthesis need chlorophyll?'' would
be erroneously penalized.

\rev{Training uses this reward in a two-stage pipeline.
We first warm up Qwen2.5-7B-Instruct with supervised fine-tuning on the
797-trajectory {\dataname} corpus to establish basic Socratic response patterns.
Starting from the SFT checkpoint, the second stage applies GRPO: for each prompt,
multiple tutor continuations are sampled, scored by $\rtotal$, normalized within
the group to obtain relative advantages, and used for policy updates against a
frozen reference model.}

\section{Evaluation Metrics}
\label{sec:eval}

\subsection{Scaffolding Effectiveness (\SE{})}
\rev{Following outcome-oriented tutoring evaluation
\cite{ilkou2025hybrid,liu2025guideeval}, {\SE} measures the proportion of test
questions where the tutor guides a simulated student to self-express the core
concept \emph{without} uttering any term in $\mathcal{K}_i$:}
\begin{equation}
  \SE = \frac{1}{N}\sum_{i=1}^{N}
    \mathbb{1}\!\left[
      \hat{c}_i = c_i^* \;\land\;
      \nexists\,t\leq T: k\in r_t,\; k\in\mathcal{K}_i
    \right]
\end{equation}
\rev{The tutor and simulator interact for up to $T{=}4$ rounds; a separate judge
then checks final concept match and tutor-side keyword leakage.}

\subsection{Conversation Depth (\CD{})}
\rev{Following progression-aware dialogue evaluation~\cite{sanders2022progression},
{\CD} measures the fraction of consecutive guide--student turn pairs
$(r_t, q_{t+1})$ whose Sentence-BERT embeddings~\sbertcite{} have cosine
similarity $<0.70$, indicating semantic progression.
We report {\CD} with {\SE} and leakage because novelty alone does not guarantee
successful scaffolding.}

\subsection{Simulator Validity}
\rev{Because {\SE} and {\CD} are simulation-based, we validate the student proxy
against held-out real follow-ups~\cite{scarlatos2026simulated}.
Across 30 questions with recorded follow-ups, each simulated utterance is matched
to the nearest real child message from the same seed by SBERT similarity; the
mean score is $0.449 \pm 0.220$, indicating on-topic but non-duplicate behavior.}

\section{Experiments}
\label{sec:exp}

We evaluate whether GRPO improves scaffolding beyond SFT and prompting
(RQ1), which reward components drive {\SE} and leakage (RQ2), and whether
scale can substitute for task-specific alignment (RQ3).

\subsection{Experimental Setup}
\label{sec:exp_setup}

\paragraph{Compared Systems.}
We compare a \emph{parameter-training (PT) track} on a fixed 7B backbone with a
\emph{prompt-engineering (PE) reference track}:
\begin{itemize}[leftmargin=1.5em,noitemsep]
\item \textbf{PT-Base}: Qwen2.5-7B-Instruct with a Socratic system
  prompt; no gradient updates.
\item \textbf{PT-SFT}: SFT on 797 trajectories only (Phase~1).
\item \textbf{PT-GRPO Full}: Phase~1 $+$ Phase~2 with
  $\alpha{=}\beta{=}0.4$, $\gamma{=}0.2$.
\item \textbf{PT-GRPO w/o $\rdir$}: Phase~1 $+$ Phase~2 with
  $\alpha{=}\beta{=}0.4$, $\gamma{=}0$.
\end{itemize}
We also include \textbf{Qwen-72B-Direct} to test scale and \textbf{PE-GRPO}
(Qwen-plus with a Socratic prompt and term constraints) as a prompting reference.

\paragraph{Implementation Details.}
Table~\ref{tab:impl} summarizes shared settings.
PT variants use the same initialization, chat template, and {\dataname} corpus;
only Phase~2 reward weights differ.
The simulator and judge are fixed across systems so score differences reflect
tutor behavior rather than evaluation noise.

\begin{table}[H]
\caption{Training and evaluation hyperparameters.}
\label{tab:impl}
\centering\small
\begin{tabular}{ll}
\toprule
\textbf{Setting} & \textbf{Value} \\
\midrule
\multicolumn{2}{l}{\textit{Phase 1 (SFT)}} \\
Base model & Qwen2.5-7B-Instruct \\
Training data & 797 {\dataname} trajectories \\
Epochs / batch size & 3 / 16 \\
\midrule
\multicolumn{2}{l}{\textit{Phase 2 (GRPO)}} \\
Group size $G$ & 8 \\
PPO clip $\varepsilon$ / KL coeff.\ $\lambda$ & 0.2 / 0.01 \\
Epochs / batch size & 1 / 16 \\
Reward weights (default) & $\alpha{=}\beta{=}0.4$; $\gamma{=}0.2$ or $0$ \\
Hardware & 1$\times$ NVIDIA H800 (80\,GB) \\
\midrule
\multicolumn{2}{l}{\textit{Evaluation protocol}} \\
Test questions $N$ & 30 held-out seeds \\
Max dialogue turns $T$ & 4 \\
Student simulator / concept judge & Qwen-plus / Qwen-turbo \\
{\CD} encoder & Sentence-BERT~\sbertcite{} ($\theta{=}0.70$) \\
\bottomrule
\end{tabular}
\end{table}

\paragraph{Test Set.}
We use 30 held-out seed questions stratified across the seven clustered science
domains.  Seeds are excluded from training trajectories and synthetic replay
prompts, and each item has a fixed gold concept $c_i^*$ plus keyword set
$\mathcal{K}_i$ for leakage detection.

\paragraph{Metrics and Model Selection.}
We report {\SE}, leakage, {\CD}, and
$J=\SE-0.5\times\text{LeakRate}$, which penalizes success achieved by keyword
leakage.
Because $N{=}30$ gives per-run {\SE} variance of roughly $\pm0.10$, margins
below 0.10 are treated as indicative; a reward-weight sensitivity check is
summarized in
Appendix~\ref{app:hpo}.

\section{Results and Analysis}
\label{sec:results}

\subsection{Main Results (RQ1)}

Table~\ref{tab:main} compares all systems on the 30-question test set.
Within the PT track, alignment yields a generally improving pattern:
prompt-only PT-Base reaches {\SE}${=}0.300$;
SFT alone improves to 0.500;
GRPO without $\rdir$ further raises {\SE} to 0.633 while cutting leakage
from 26.7\% to 13.3\%.
PT-GRPO w/o $\rdir$ also achieves the highest {\CD} (0.722), suggesting
that stronger scaffolding is accompanied by deeper semantic progression rather
than simple paraphrase.
Relative to PE-GRPO, the best 7B PT model is higher in {\SE}
($0.633$ vs.\ $0.400$) and $J$ ($0.567$ vs.\ $0.367$), despite using a
smaller backbone. This supports gradient-based alignment over prompting alone
for this behavioral objective.

\begin{table}[H]
\caption{Results on 30 held-out questions.
  $J = \SE - 0.5 \times \text{LeakRate}$.
  {\CD} uses SBERT cosine similarity.
  \textbf{Bold} = best PT-track value.
  $\dagger$~PE-GRPO is a prompt-engineering reference, not a direct PT
  competitor.}
\label{tab:main}
\centering\small
\begin{tabular}{llcccc}
\toprule
\textbf{Category} & \textbf{System} &
  \textbf{SE}$\uparrow$ & \textbf{LeakRate}$\downarrow$ &
  \textbf{CD}$\uparrow$ & \textbf{J}$\uparrow$ \\
\midrule
\multirow{2}{*}{\shortstack[l]{Scale\\Baseline}}
  & Qwen-72B-Direct      & 0.000 & 0.967 & --- & $-$0.483 \\
  & PT-Base (7B, prompt) & 0.300 & 0.300 & 0.667 & \phantom{$-$}0.150 \\
\midrule
\multirow{3}{*}{\shortstack[l]{PT Track}}
  & PT-SFT
    & 0.500 & 0.267 & 0.689 & \phantom{$-$}0.367 \\
  & PT-GRPO Full ($\alpha{=}\beta{=}0.4,\,\gamma{=}0.2$)
    & 0.433 & 0.233 & 0.644 & \phantom{$-$}0.317 \\
  & \textbf{PT-GRPO w/o $\rdir$} ($\gamma{=}0$)
    & \textbf{0.633} & \textbf{0.133} & \textbf{0.722} &
      \textbf{\phantom{$-$}0.567} \\
\midrule
PE Ref.$^\dagger$
  & PE-GRPO
    & 0.400 & 0.067 & --- & \phantom{$-$}0.367 \\
\bottomrule
\end{tabular}
\end{table}

\begin{figure}[H]
\centering
\begin{minipage}[b]{0.46\textwidth}
\centering
\begin{tikzpicture}
\begin{axis}[
  ybar, bar width=10pt,
  width=0.95\linewidth, height=52mm,
  title={\small\textbf{(a) Scaffolding Effectiveness (SE)}},
  ylabel={\small SE},
  symbolic x coords={Base,SFT,Full,w/o Rdir},
  xtick=data,
  xticklabels={Base,SFT,Full,w/o $R_{dir}$},
  xticklabel style={font=\scriptsize, align=center},
  ymin=0, ymax=0.85,
  ytick={0,0.2,0.4,0.6,0.8},
  yticklabel style={font=\scriptsize},
  title style={font=\small},
  ylabel style={font=\scriptsize},
  axis lines=left,
  enlarge x limits=0.2,
  nodes near coords,
  nodes near coords style={font=\scriptsize},
]
\addplot[fill=blue!35, draw=blue!60]
  coordinates {(Base,0.300)(SFT,0.500)(Full,0.433)(w/o Rdir,0.633)};
\end{axis}
\end{tikzpicture}
\end{minipage}
\hfill
\begin{minipage}[b]{0.46\textwidth}
\centering
\begin{tikzpicture}
\begin{axis}[
  ybar, bar width=10pt,
  width=0.95\linewidth, height=52mm,
  title={\small\textbf{(b) Keyword Leakage Rate}},
  ylabel={\small LeakRate},
  symbolic x coords={Base,SFT,Full,w/o Rdir},
  xtick=data,
  xticklabels={Base,SFT,Full,w/o $R_{dir}$},
  xticklabel style={font=\scriptsize, align=center},
  ymin=0, ymax=0.55,
  ytick={0,0.1,0.2,0.3,0.4,0.5},
  yticklabel style={font=\scriptsize},
  title style={font=\small},
  ylabel style={font=\scriptsize},
  axis lines=left,
  enlarge x limits=0.2,
  nodes near coords,
  nodes near coords style={font=\scriptsize},
]
\addplot[fill=red!30, draw=red!60]
  coordinates {(Base,0.300)(SFT,0.267)(Full,0.233)(w/o Rdir,0.133)};
\end{axis}
\end{tikzpicture}
\end{minipage}
\caption{PT-track {\SE} and leakage on the 30-question test set.
  PT-GRPO w/o $\rdir$ achieves the best performance on both metrics;
  adding $\rdir$ during GRPO lowers {\SE} despite reducing leakage.}
\label{fig:bars}
\end{figure}
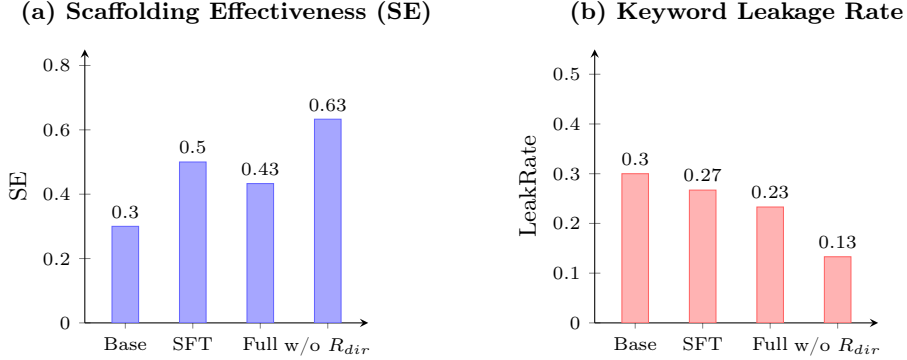

Figure~\ref{fig:bars} visualizes the same PT-track progression.
{\SE} rises from PT-Base through SFT and peaks at PT-GRPO w/o $\rdir$;
leakage declines over the same sequence.
The dip at PT-GRPO Full (middle bars) foreshadows the reward-conflict
analysis in Section~\ref{sec:ablation}.

\subsection{Reward Ablation (RQ2)}
\label{sec:ablation}

To isolate reward effects (RQ2), we compare PT-track variants with
components selectively enabled or disabled.
Table~\ref{tab:ablation} summarizes the PT grid.
On the PT track, removing $\rdir$ from optimization strictly improves
both {\SE} and leakage over the full reward.
This contrasts with the PE-GRPO reference in Table~\ref{tab:main}, where
term constraints are applied only at decoding time; the result suggests that
$\rdir$ is risky as a gradient signal even when explicit constraints can reduce
inference-time leakage.

\begin{table}[H]
\caption{Reward component ablation on the PT track ($n{=}30$).
  Removing $\rdir$ strictly improves both metrics: SE increases by
  0.200 and leakage falls by 10 percentage points relative to the
  full reward.
  ``\ding{51}'' = component active; ``\ding{55}'' = disabled.}
\label{tab:ablation}
\centering\small
\begin{tabular}{lccccc}
\toprule
\textbf{Variant} & $\rcog$ & $\reng$ & $\rdir$ &
  \textbf{SE}$\uparrow$ & \textbf{LeakRate}$\downarrow$ \\
\midrule
PT-Base (no GRPO)
  & \ding{55} & \ding{55} & \ding{55} & 0.300 & 0.300 \\
PT-SFT
  & \ding{55} & \ding{55} & \ding{55} & 0.500 & 0.267 \\
PT-GRPO w/o $\rdir$
  & \ding{51} & \ding{51} & \ding{55} & \textbf{0.633} & \textbf{0.133} \\
PT-GRPO Full
  & \ding{51} & \ding{51} & \ding{51} & 0.433 & 0.233 \\
\bottomrule
\end{tabular}
\end{table}

\subsection{Qualitative Analysis}
\label{sec:qualitative}

Table~\ref{tab:qual} illustrates the behavioral shift induced by
GRPO alignment.

\begin{table}[H]
\caption{Qualitative comparison of Base vs.\ \modelname{}-GRPO responses.
  Underlined terms are core vocabulary from $\mathcal{K}_i$.
  The GRPO-aligned model consistently opens with analogies and exploratory
  questions rather than direct disclosure.}
\label{tab:qual}
\centering\small
\begin{tabular}{@{}p{0.24\linewidth}p{0.34\linewidth}p{0.34\linewidth}@{}}
\toprule
\textbf{Student Question} & \textbf{Base (Direct)} & \textbf{\modelname{}-GRPO} \\
\midrule
``Why is the sky blue?'' &
``The sky appears blue because of \underline{Rayleigh scattering}.
Shorter wavelength (blue) light scatters more in Earth's atmosphere.'' &
``Great question! Have you noticed that the sky looks different at sunset?
What color is it then, and why do you think that might change? $\boldsymbol{\to}$'' \\
\midrule
``Why do plants need sunlight?'' &
``Plants need sunlight for \underline{photosynthesis}, where
\underline{chlorophyll} converts CO$_2$ and water into glucose.'' &
``Imagine a plant in complete darkness for a week---what do you think would
happen to it? What do you think sunlight gives the plant that it can't get
elsewhere? $\boldsymbol{\to}$'' \\
\bottomrule
\end{tabular}
\end{table}

\subsection{Discussion of Findings}

\paragraph{Training trajectory (RQ1).}
The PT track follows an alignment ladder with one important caveat
(Table~\ref{tab:main}, Figure~\ref{fig:bars}).
SFT provides the largest single-step gain in {\SE} (+0.200 over
PT-Base) by instilling basic questioning patterns from {\dataname},
but it still leaks target terms in 26.7\% of dialogues.
GRPO with $\rcog$ and $\reng$ further reshapes turn-level behavior:
the w/o-$\rdir$ variant improves {\SE} by another 0.133 points and
halves leakage relative to SFT.
Notably, PT-GRPO Full \emph{regresses} relative to PT-SFT on {\SE},
showing that reward design matters as much as the choice to apply RL at
all.

\paragraph{Parameter training vs.\ prompt engineering (RQ1).}
PT-GRPO w/o $\rdir$ outperforms PE-GRPO in \SE{}
($0.633$ vs.\ $0.400$) and $J$ ($0.567$ vs.\ $0.367$), while using a
7B backbone rather than an API model.
Gradient-based alignment directly updates the model's conditional
output distribution, whereas prompt engineering only influences the
decoding context at inference time; this structural difference
explains the persistent performance gap even with a smaller base
model.
PE-GRPO achieves lower leakage (6.7\%) via explicit term constraints,
but cannot close the {\SE} gap without parameter updates.

\paragraph{The counterproductive role of $\rdir$ (RQ2).}
Adding $\rdir$ with $\gamma{=}0.2$ reduces \SE{} by 0.200 and raises
leakage by 10 percentage points relative to $\gamma{=}0$.
We attribute this to an optimization conflict: $\rdir$ applies a
gradient signal that suppresses high-frequency scientific vocabulary
broadly, which in turn reduces the model's ability to use precise
terminology as scaffolding cues.
Because $\reng$ rewards engagement through linguistic richness,
the two objectives partially cancel when $\gamma > 0$, leaving
a net degradation.
This interaction is consistent with the 9-trial Bayesian hyperparameter
search (Appendix~\ref{app:hpo}): all configurations with
$\gamma \geq 0.20$ and $\beta \leq 0.40$ yield $J < 0$.
The small grid therefore favors low $\gamma$ and sufficiently large
$\beta$, although larger validation sets are needed to draw a sharp boundary.

\paragraph{Scale does not induce Socratic behavior (RQ3).}
Qwen-72B without task-specific alignment achieves 0\% \SE{} and 96.7\%
leakage, matching the direct-answering failure observed in the unaligned
API baseline rather than the prompted PT-Base system.
Instruction-following capability, even at large scale, does not
produce the behavioral restraint required for Socratic tutoring;
explicit optimization against a pedagogical objective is necessary.

\section{Conclusion}
\label{sec:conclusion}

We presented \modelname{}, a training pipeline for redirecting a
7B instruction-tuned LLM from direct answer delivery toward Socratic
guided inquiry.
The approach combines a 797-trajectory Chinese science dialogue corpus
(\dataname{}), a three-component heuristic reward with a
student-vocabulary correction ($\kquery$), and two outcome-oriented
evaluation metrics (\SE{}, \CD{}) assessed through an LLM student
simulator.

Three substantive findings emerge from controlled experiments.
First, GRPO parameter training yields stronger Socratic behavior than
prompt engineering in our setting: the 7B PT-GRPO model achieves
SE\,=\,0.633 versus 0.400 for the PE-GRPO reference, indicating that
gradient-level alignment can be more effective than inference-time prompting
for this objective.
Second, explicit term constraints can reduce inference-time leakage, but
using the same directness signal as a GRPO reward is harmful in parameter
training, where its interaction with the engagement reward $\reng$ produces
a net degradation in both \SE{} and leakage.
Third, unaligned instruction-following at 72B scale is no better
than a 7B baseline on this task, underscoring that Socratic
alignment is a targeted behavioral problem requiring explicit
optimization rather than a byproduct of general model capability.

\bibliographystyle{splncs04}
\bibliography{references}

\appendix

\section{Reward Weight Sensitivity}
\label{app:hpo}

We run a nine-trial Optuna TPE search over
$\alpha \in [0.10,0.70]$, $\beta \in [0.20,0.80]$, and
$\gamma \in [0.00,0.30]$, optimizing
$J=\SE-0.5\times\text{LeakRate}$.
The best searched configuration
($\alpha{=}0.45,\beta{=}0.80,\gamma{=}0.10$) obtains
$\SE{=}0.400$ and $J{=}0.217$, below the $\gamma{=}0$ setting in
Table~\ref{tab:main}; all trials with $\gamma \geq 0.20$ and
$\beta \leq 0.40$ yield $J<0$.
Thus the search is used only as sensitivity evidence that high engagement
weight is helpful and larger directness penalties are unstable.

\begin{credits}
\subsubsection{\ackname}
We thank the partner schools and the Dr.\ Curious platform team for
data access, and the China Association of STEM Education for research
support.

\subsubsection{\discintname}
The authors have no competing interests to declare.
\end{credits}

\end{document}